\documentclass[10pt, a4paper]{article}
\usepackage{lrec}
\usepackage{graphicx}
\usepackage{xcolor}
\usepackage{tabularx}
\usepackage{soul}
\usepackage{dcolumn}
\usepackage[latin1]{inputenc}
\usepackage{combelow}
\usepackage{hyperref}
\usepackage{microtype}
\usepackage{url}
\usepackage{booktabs}
\usepackage{xspace}
\newcommand{\eg}{\textit{e.\,g.}\xspace}

\newcommand{\F}{$\textrm{F}_1$\xspace}

\title{\textbf{Automatic Section Recognition in Obituaries}}

\name{Valentino Sabbatino, Laura Bostan, Roman Klinger}

\address{University of Stuttgart\\ Institut f\"ur Maschinelle Sprachverarbeitung \\
  Pfaffenwaldring 5b, 70569 Stuttgart, Germany \\
  \{valentino.sabbatino,laura.bostan,roman.klinger\}@ims.uni-stuttgart.de\\}

\abstract{Obituaries contain information about people's values across
  times and cultures, which makes them a useful resource for exploring
  cultural history. They are typically structured similarly, with
  sections corresponding to \textit{Personal Information},
  \textit{Biographical Sketch}, \textit{Characteristics},
  \textit{Family}, \textit{Gratitude}, \textit{Tribute},
  \textit{Funeral Information} and \textit{Other} aspects of the
  person. To make this information available for further studies, we
  propose a statistical model which recognizes these sections. To
  achieve that, we collect a corpus of $20058$ English obituaries from
  \href{http://obituaries.dailyitem.com/obituaries/all-categories/search/}{\emph{TheDaily
      Item}},
  \href{http://www.remembering.ca/obituaries/all-categories/search/}{\emph{Remembering.CA}}
  and \href{http://lfpress.remembering.ca/}{\emph{The London Free
      Press}}. The evaluation of our annotation guidelines with three
  annotators on 1008 obituaries shows a substantial agreement of
  Fleiss $\kappa=0.87$. Formulated as an automatic segmentation task,
  a convolutional neural network outperforms bag-of-words and
  embedding-based BiLSTMs and BiLSTM-CRFs with a micro $F_1=0.81$.\\
  \newline 
}

\begin{document}

\maketitleabstract
\section{Introduction and Motivation}

An obituary, typically found in newspapers, informs about the recent death of a
person, and usually includes a brief bio\-graphy of the deceased person, which
sometimes recounts detailed life stories and anecdotes. Structural elements,
styles, formats, and information presented vary slightly from culture to
culture or from community to community~\cite{moses2003obituaries}. Obituaries
can be considered to be short essays and contain information on the living
family members and information about the upcoming funeral, such as visitation,
burial service, and memorial information as well as the cause of death
\cite{moses2003obituaries}.

\begin{table*}
  \begin{tabularx}{\linewidth}{p{2cm}Xp{3cm}}
    \toprule
    Zone & Text & Relevant Elements \\
    \cmidrule(r){1-1}\cmidrule(rl){2-2}\cmidrule(l){3-3}
    Personal\par Information &
    John of Bad Cannstatt, passed away peacefully on
    November 23, 2018 at the age of 52. &
    Name, Location, Mode of Death, Date of Death, Age\\
    \cmidrule(r){1-1}\cmidrule(rl){2-2}\cmidrule(l){3-3}
    Family & John will be lovengly remembered by his
    children Mary and Laura, his parents Valentino and Nora, brother Andrew
    (Karolin), Jason and niece, Sebastian plus many friends. &
    Mentions of Children, Parents, Family, Friends\\
    \cmidrule(r){1-1}\cmidrule(rl){2-2}\cmidrule(l){3-3}
    Characteristics & John loved gold, hockey, football, water
    skiing, downhill skiing spending time with his kids and coaching
    their ringette and hockey teams. &
    Hobbies, Interests\\
    \cmidrule(r){1-1}\cmidrule(rl){2-2}\cmidrule(l){3-3}
    Funeral\par Information  & Monday, December 3 at Cannstatt Church, 9009 163 St SW, Stuttgart, XYZ at 10:00am. &
    Date, Location, Time \\
    \bottomrule
  \end{tabularx}
  \caption{Example of an annotated obituary together with specific constituent elements for each of the zones.}
  \label{tab:example}
\end{table*}

Similarly to biographies, obituaries represent an interesting type of text
because the information contained is usually focused on the values and the
qualities of a given human being that is part of a particular community
\cite{Hume2000,Kinnier1994,Long1987}. From the digital humanities perspective
investigating obituaries also provides an understanding of how the community
who writes the obituaries decides what is relevant about life and death.

Potential applications that are enabled by having access to large collections
of obituaries are finding such themes that are relevant while discussing life
and death, investigation of different aspects of social memory
\cite{Fowler2011,Arnason2003} (finding what is being remembered or chosen to be
excluded from an obiturary), investigation of correlations between work or
other different themes and the cause of death, analysis of linguistic,
structural or cultural differences \cite{Bytheway1996}, and the investigation
of different biases and values within a community
\cite{Simoni1983,Hume2003,David2002,Chang2018}.

More recently, obituaries have been published on dedicated social networks
where the mourners who write the obituaries express their emotions and tell
stories of the deceased in comments to the obituaries (\eg Legacy.com,
Remembering.CA). These networks facilitate interactions between readers and the
family of the deceased \cite{Hume2010}. With this paper, we focus on online
publications of obituaries which are available online and are in English.

Research that builds on top of such data is presumably mostly concerned with a
part of the information contained in obituaries. For example, when
investigating mortality records \cite{soowamber2016}, one might only be
interested in the \textit{Personal Information} section. Therefore, we propose
to perform zoning as a preprocessing step and publish a corpus and trained
models for the sections \textit{Personal information} (including names of the
deceased, birth date, date of death, and cause of death), \textit{Biographical
sketch}, \textit{Tribute}, \textit{Family}, and \textit{Funeral Information}
(such as time, place, and date of the funeral). No such resource is currently
available to the research community.

Our main contributions are therefore (1) to annotate a collection of
obituaries, (2) to analyze the corpus and to formulate the task of
automatic recognition of structures, (3) to evaluate which models
perform best on this task, and (4) to compare the models' results
qualitatively and quantitatively.  To achieve our goals and as
additional support for future research, we publish information how to
obtain the data and the annotated dataset as well as the models at
\url{http://www.ims.uni-stuttgart.de/data/obituaries}.

\section{Related Work}
Research on obituaries can be structured by research area, namely
language studies, cultural studies, computational linguistics,
psychology studies, and medical studies.

\subsection{Obituaries in Cultural and Medical Studies}
One of the common topics that are  studied in the context of cultural studies and
obituaries is religion. \newcite{herat2014} investigate how certain language
expressions are used in obituaries in Sri Lanka, how religion and culture play
a role in the conceptualization of death, and how language reflects social
status. They find that the conceptualization of death is in terms of a journey
in the Buddhist and Hindu communities whereas death is conceptualized as an end
in Christian and Muslim communities. They show that the language of obituaries
appears to be conditioned by the religious and cultural identity of the
deceased.

\newcite{ergin2012} look into Turkish obituary data from \textit{H\"urriyet}, a
major Turkish daily newspaper, from 1970 to 2009, with the goal of finding
expressions of religiosity and constructions of death in relation to gender and
temporal variations together with markers of status. Their results show that
the obituaries considered are relying on ``an emotional tone of loss'' and that
the spiritual preferences are linked to the status and appartenance to a
specific social class.

Next to religion, elements of the obituary language are in the focus of various
works across countries and cultures. \newcite{metaphors2019} undertake a
qualitative analysis of metaphors in 150 obituaries of professional athletes
published in various newspapers. They find traditional metaphors of death but
also creative metaphors that describe death euphemistically. Some of the
creative metaphors have a connection to sports but not necessarily to the sport
practiced by the deceased athlete.

The language of obituaries is also investigated in the context of gender
analysis by \newcite{malesvsfemales} who test the hypothesis that obituaries
are less emotional in the language used for females than for males. They
collect 703 obituaries from a local newspaper from US and investigate whether
the person is described to have ``died'' or ``passed away''. Their results show
that the deaths of females are more likely to be described as ``passing away''.

Furthermore, the perception of women in leading positions in communist and
post-communist Romania is researched by \newcite{gender2011} by analyzing the
content of obituaries published in the Romanian newspaper \textit{Rom\^{a}nia
Liber\u{a}} from 1975 to 2003. They show that the gender gap in management
widened after the fall of communism.

\newcite{epstein2013} study the relationship between career success, terminal
disease frequency, and longevity using New York Times obituaries. Their results
show that obituaries written in the memory of men are more prevalent and the
mean age of death was higher for males than females. They concluded that
``smoking and other risk behaviours may be either the causes or effects of
success and/or early death'', and fame and achievement in performance-related
careers correlate with a shorter life span expectancy.

\newcite{rusu2017} also look at famous people, and the posthumous articles
written about them to test whether the deceased are protected from negative
evaluations within their community. They find out that more than one fifth of
the articles do contain negative evaluations of the deceased.

\newcite{barth2013} gains insights into how different communities deal with
death according to their respective norms. They study the differences between
German and Dutch obituaries in terms of visual and textual elements,
information about the deceased, and funeral-related information. Their study
shows that German obituaries use illustrations more than the Dutch ones and
that the Dutch obituaries provide more information than the German ones.

Another cross-cultural study is made by \newcite{hubbard2009} who investigate
whether obituaries placed by families reflect specific societal attitudes
towards aging and dementia. They use discourse analysis of obituaries in
newspapers from Canada and the UK and show that donations to dementia charities
were more common in obituaries from Canada than in the UK.

\newcite{themes_opiod} study the public perception on the opioid epidemic in
obituaries from the US where the cause of death is related to overdose. They
investigated emotion related themes and categories by using the IBM Watson Tone
Analyzer\footnote{\url{https://www.ibm.com/watson/services/tone-analyzer/},
last accessed \today} and show that joy and sadness are the most prevalent
emotion categories with the most common emotion being love. The terms that are
most used to describe death are ``accidental'' and ``addiction''. Shame and
stigma are less prevalent ``which might suggest that addiction is perceived as
a disease rather than a criminal behaviour''.

\newcite{usobi} investigate the shared values of the community of neurosurgeons
in the US by doing a text analysis on obituaries from Neurosurgery, Journal of
Neurosurgery and the New York Times. Their study analyzes frequent terms and
derives the relative importance of various concepts: innovation, research,
training and family. Within this work, the sentiment of the obituaries within
the Neurosurgery research community is being annotated. A result of this study
is that the obituaries of neurosurgeons written by the research community put a
greater emphasis on professional leadership and residency training and that the
family mentions occured more in the lay press.

\newcite{vital} develop a methodology to link mortality data from internet
sources with administrative data from electronic health records. To do so they
implement and evaluate the performance of different linkage methods. The
electronic health records are from patients in Rennes, France and the extracted
obituaries are all available online obituaries from French funeral home
websites. They evaluate three different linkage methods and obtain almost
perfect precisions with all methods. They conclude that using obituaries
published online could address the problem of long delays in the sharing of
mortality data whereas online obituaries could be considered as reliable
data source for real-time suveillance of mortality in patients with cancer.

\subsection{Obituaries as a Data Source in Various Tasks of Computational Linguistics}

With a focus on computational linguistics, \newcite{obituary_mining1} analyze
text data from obituary websites, with the intention to use it to prevent
identity theft. The goal was to evaluate how ``often and how accurately name
and address fragments extracted from these notices developed into complete name
and address information corresponding to the deceased individual''. They use a
knowledge base with name and address information, extracte the name and
address fragments from the text and match them against the knowledge base to
create a set of name and address candidates. This result set is then compared
to an authoritative source in order to determine which of the candidate records
actually correspond to the name and address of an individual reported as
deceased.

\newcite{alfano2018} collect obituaries from various newspapers, to get a
better understanding of people's values. They conduct three studies in which
the obituaries are annotated with age at death, gender and general categories
that summarize traits of the deceased (a trait like hiker would be summarized
by the category ``nature-lover''). All studies are analyzed from a network
perspective: when the deceased is described as having the traits X and Y, then
an edge between the two traits is created with the weight of the edge being the
total number of persons described as having both traits. The first study is
done on obituaries collected from local newspapers. They find that women's
obituaries focus more on family and ``care-related affairs'' in contrast to
men's obituaries which focus on ``public and political matters''. In the second
study they explore the New York Times Obituaries and find that the network of
the second study differs from the first study in terms of network density, mean
clustering coefficient and modularity. The last study is done on data from
ObituaryData.com and the annotation with traits is performed in a
semi-automatic manner.

\newcite{obi1} extract various facts about persons from obituaries. They use a
feature scoring method that uses prior knowledge. Their method achieved high
performance for the attributes person name, affiliation, position (occupation),
age, gender, and cause of death.

\newcite{bamman2014} present an unsupervised model for learning life event
classes from biographical texts in Wikipedia along with the structure that
connects them. They discover evidence of systematic bias in the presentation
of male and female biographies in which female biographies placed a
significantly disproportionate emphasis on the personal events of marriage and
divorce. This work is of interest here because it handled biographical
information (Wikipedia biographies), of which obituaries are also a part.

\newcite{simonson2016} investigate the distribution of \textit{narrative
schemas}~\cite{Chambers2009} throughout different categories of documents and
show that the structure of the narrative schemas are conditioned by the type of
document. Their work uses the New York Times corpus, which makes the work
relevant for us, because obituary data is part of the NYT library and a
category of document the work focuses on. Their results show that obituaries
are narratologically homogeneous and therefore more rigid in their wording and
the events they describe.

The stability of narrative schemas is explored in a follow up paper by
\newcite{simonson2018}. Their goal was to test whether small changes in the
corpus would produce small changes in the induced schemas. The results confirm
the distinction between the homogeneous and heterogeneous articles and show
that homogeneous categories produced more stable batches of schemas than the
heterogeneous ones. This is not surprising but supports that obituaries have a
coherent structure which could be turned into a stable narrative schema.

\newcite{he2019} propose using online obituaries as a new data source for doing
named entity recognition and relation extraction to capture kinship and family
relation information. Their corpus consists of 1809 obituaries annotated with a
novel tagging scheme. Using a joint neural model they classify to 57 kinships
each with 10 or more examples in 10-fold cross-validation experiment.

\begin{table}
  \centering
  \begin{tabular}{llr}
    \toprule
    Source & Location & \# obituaries\\
    \cmidrule(r){1-1}\cmidrule(r){2-2}\cmidrule(l){3-3}
    Daily Item & US & 9975\\
    Remembering CA & Canada & 9984\\
    The London Free Press & UK & 99\\
   \cmidrule(r){1-1}\cmidrule(r){2-2}\cmidrule(l){3-3}
    All & All & 20058  \\
    \bottomrule
  \end{tabular}
  \caption{Overview of the sources of obituary data.}
  \label{crawled}
\end{table}

\subsection{Zoning}

Many NLP tasks focus on the extraction and abstraction of specific types of
information in documents. To make searching and retrieving information in
documents accessible, the logical structure of documents in titles, headings,
sections, arguments, and thematically related parts must be
recognized~\cite{structuring}.

A notable amount of work focuses on the
\textit{argumentative zoning} of scientific documents
\cite{teufel1999,teufel2002,teufel2009,liakata2010,contractor2012,ravenscroft2016,neves2019}.
\newcite{zoning2} stated that readers of scientific work may be looking for
``information about the objective of the study in question, the methods used in
the study, the results obtained, or the conclusions drawn by authors''.

The recognition of document structures generally makes use of two sources of
information. On one side, text layout enables recognition of relationships
between the various structural units such as headings, body text, references,
figures, etc. On the other side, the wording and content itself can be used to
recognize the connections and semantics of text passages. Most methods use
section names, argumentative zoning, qualitative dimensions, or the conceptual
structure of documents \cite{zoning2}.

Common to all the works that focus on zoning of scientific articles is
the formulation or use of an annotation scheme, which in this case relies on
the form and meaning of the argumentative aspects found in text rather than on
the layout or contents. In contrast to argumentative zoning, our work does not
make use of an annotation scheme of categories that relate to rhetorical
moves of argumentation \cite{teufel1999}, but focuses instead on content.

\section{Data}
\label{Dataset}

\subsection{Collection}

We collected obituaries from three websites: \emph{The Daily
  Item}\footnote{\url{http://obituaries.dailyitem.com/obituaries/all-categories/search/}},
where obituaries from the USA are published,
\emph{Remembering.CA}\footnote{\url{http://www.remembering.ca/obituaries/all-categories/search/}},
which covers obituaries from Canada, and \emph{The London Free
  Press}\footnote{\url{http://lfpress.remembering.ca/}}, which covers
obituaries from London (see Table~\ref{crawled}). The obituaries on
\emph{The Daily Item} and \emph{The London Free Press} are dedicated websites where
people could publish their obituaries. \emph{Remembering.CA} is an aggregator
and shows obituaries published from different sources.  The total set
consists of 20058 obituaries.

\subsection{Annotation Scheme and Guidelines}

In each obituary, we can find certain recurring elements, some
factual, such as the statement that announces the death which contains
the names of the deceased, age, date of death, information about
career, information about the context and the cause of death (detailed
if the person was young or suffering of a specific disease). The life
events and career steps are sketched after that.  This is usually
followed by a list of hobbies and interests paired with
accomplishments and expressions of gratitude or a tribute from the
community of the deceased. Towards the end of the obituary, there are
mentions of family members (through names and type of relation). The
obituaries commonly end with details about the
funeral~\cite{moses2003obituaries}.

Therefore, we define the following eight classes: \textit{Personal
  information}, \textit{Biographical sketch},
\textit{Characteristics}, \textit{Tribute}, \textit{Expression of
  gratitude}, \textit{Family}, \textit{Funeral information}, and
\textit{Other} to structure obituaries at the sentence level. An
example of these classes in context of one obituary is depicted in
Table~\ref{tab:example}.

The \textit{Personal Information} class serves the purpose to classify
most of the introductory clauses in obituaries. We have chosen to
refer to a sentence as \textit{Personal Information} when it includes
the name of the deceased, the date of death, the cause of death, or
the place of death. For example \textit{John Doe, 64, of Newport,
  found eternal rest on Nov. 22, 2018.}

The \textit{Biographical sketch} is similar to a curriculum
vitae. Sections in a person's life fall into this category. However,
it should not be regarded exclusively as a curriculum vitae, since it
forms the superset of personal information. We decided to label a
sentence as \textit{Biographical sketch} if it includes the place of
birth, the date of birth, the last place of residence, the wedding
date, the duration of the marriage, the attended schools, the
occupations, or the further events in life. An example is \textit{He
  entered Bloomsburg State Teachers College in 1955 and graduated in
  1959.}

The class \textit{Characteristics} is recognizable by the fact that the
deceased person is described through character traits or things the dead person
loved to do. Apart from hobbies and interests, the deceased's beliefs are also
part of the characteristics. An example is \textit{He enjoyed playing
basketball, tennis, golf and Lyon's softball.}

Sentences about major achievements and contributions to society are
labeled as \textit{Tribute}. An example is \textit{His work was a
  credit to the Ukrainian community, elevating the efforts of its arts
  sector beyond its own expectations.}

Sentences in obituaries are labeled as an expression of \textit{Gratitude} if
any form of gratitude occurs in it, be it directed to doctors, friends, or
other people. In most cases, it comes from the deceased's family. An example
is \textit{We like to thank Leamington Hospital ICU staff,
Windsor Regional Hospital ICU staff and Trillium for all your great care and
support.}

The class \textit{Family} is assigned to all sentences that address the
survivors or in which previously deceased close relatives, such as siblings or
partners, are mentioned. The mentioning of the wedding date is not covered by
this category, because we consider it an event and as such, it falls under the
\textit{Biographical sketch} category. If the precedence of those persons
is mentioned it falls in this category. If a marriage is mentioned without the
wedding date or the duration it falls into the \textit{Family} category. An
example is: \textit{Magnus is survived by his daughter Marlene
(Dwight), son Kelvin (Patricia), brother Otto (Jean) and also by numerous
grandchildren \& great grandchildren, nieces and nephews.}

Sentences are labeled as \textit{Funeral information} when they contain
information related to the funeral, such as date of the funeral, time of the
funeral, place of the funeral, and where to make memorial contributions. An
example is \textit{A Celebration of Life will be held at the
Maple Ridge Legion 12101-224th Street, Maple Ridge Saturday December 8, 2018
from 1 to 3 p.m.}

Everything that does not fall into the above-mentioned classes is assigned the
class \textit{Other}. An example is: \textit{Dad referred to
Lynda as his Swiss Army wife.}

\subsection{Annotation Procedure and Inter-Annotator Agreement}

\begin{table}
\newcommand{\sd}{SD}
  \centering
  \begin{tabular}{lccc|c}
    \toprule
    Class &  US  &  Canada &  UK &  All  \\
   \cmidrule(r){1-1}\cmidrule(lr){2-2}\cmidrule(rl){3-3}\cmidrule(lr){4-4}\cmidrule(l){5-5}
    PI & .96 & .94 & .93 &  .94 \\
    BS & .90 & .78 & .79 &  .86\\
    FA & .93 & .89 & .90 & .91\\
    C  & .84  & .69 & .82 & .79 \\
    T & .22 & .16 & .00 & .19 \\
    G & .00 & .00 & .85 & .91 \\
    FI & .95 & .94 & .96 & .95 \\
    O & .53  & .53 & .57 & .55 \\
 \cmidrule(l){1-1}\cmidrule(l){2-2}\cmidrule(l){3-3}\cmidrule(l){4-4}\cmidrule(r){5-5}
    All & .88 & .81 & .59 & .86 \\
    \bottomrule
  \end{tabular}
  \caption{Inter-annotator agreement scores with Fleiss' $\kappa$. PI: Personal information, BS: Biographical Sketch, FA: Family, C: Characteristics, T: Tribute, G: Gratitude, FI: Funeral}
  \label{tab:agreement}
\end{table}

Our overall annotated data set consists of 1008 obituaries which are
randomly sampled from the overall crawled data. For the evaluation of
our annotation guidelines, three students of computer science at the
University of Stuttgart (all of age 23) annotate a subset of 99
obituaries from these 1008 instances. The first and second annotator
are male and the third is female. The mother tongue of the first
annotator is Italian and the mother tongue of the second and third
annotator is German. All pairwise Kappa scores as well as the overall
Fleiss' kappa scores are .87 (except for the pairwise Kappa between
the first and the second annotator, being .86). Based on this result,
the first annotator continued to label all 1008 instances.

Table~\ref{tab:agreement} reports the agreement scores by country and category.
Annotated obituaries from the UK have the lowest $\kappa=$$0.59$ and the ones
from the US the highest $\kappa=$$0.88$. Category-wise, we observed
difficulties to classify some of the rarer categories that appeared, such as
examples from the class \textit{Tribute} or \textit{Other}.   Another quite
difficult distinction is the one between the class \textit{Family} and the
class \textit{Biographical sketch} due to the occurrence of a wedding date,
which we considered an event, in connection with the other family criteria.
Furthermore we found difficult to decide on the border between \textit{Personal
Information} and \textit{Biographical sketch} zones.

\begin{table}
  \centering
  \setlength{\tabcolsep}{2pt}
  \begin{tabular}{lrrrrrr|rr}
    \toprule
    & \multicolumn{2}{c}{US: 475} & \multicolumn{2}{c}{Canada: 445} & \multicolumn{2}{c}{UK: 88} & \multicolumn{2}{|c}{All: 1008} \\
    \cmidrule(lr){2-3}\cmidrule(lr){4-5}\cmidrule(lr){6-7}\cmidrule(l){8-9}
    Class & \# sent.& \% & \# sent.& \% & \# sent.& \% & \# sent.& \% \\
   \cmidrule(r){1-1}\cmidrule(lr){2-3}\cmidrule(lr){4-5}\cmidrule(lr){6-7}\cmidrule(l){8-9}
    PI & 496 & 8 & 465 & 12 & 97 & 9 & 1058 & 10   \\
    BS & 2193 & 36  & 718 & 18 & 130 & 12 & 3041 & 27 \\
    FA & 899 & 15 & 947 & 24 & 349 & 33 & 2195 & 20 \\
    C & 787 & 13 & 375 & 10 & 72 & 7 &  1234 & 11   \\
    T & 6& 0&  5& 0& 0 & 0 & 11 & 0  \\
    G & 23 & 0 & 100 & 3 & 21 & 2 & 144 & 1  \\
    FI & 1517 & 25 & 972 & 25 & 342 & 33 & 2831 & 26  \\
    O & 177 & 3 & 363 & 9 & 33 & 3 & 573 & 5 \\
  \cmidrule(r){1-1}\cmidrule(lr){2-3}\cmidrule(lr){4-5}\cmidrule(lr){6-7}\cmidrule(l){8-9}
    All & 6098 & 100 & 3945 & 100 & 1044 & 100 & 11087 &  100  \\
    \bottomrule
  \end{tabular}
  \caption{Information on full annotated dataset of obituaries.
    PI: Personal information, BS: Biographical Sketch, FA: Family, C:
    Characteristics, T: Tribute, G: Gratitude, FI: Funeral
    Information, O: Other
 \#
    sent.\ denotes number of sentences. \% denotes the relative
counts in each class.}
  \label{tab:stat1}
\end{table}

\subsection{Analysis} \label{sec:stat}
Table~\ref{tab:stat1} shows the analysis of our 1008 annotated obituaries from
three different sources which form altogether 11087 sentences (where the
longest sentence as 321 words). 475 obituaries are from \textit{The Daily Item}
(USA), 445 obituaries are from \textit{Remembering.CA} (Canada), and 88
obituaries are from \textit{The London Free Press} (UK).  Most sentences in the
dataset are labeled as \textit{Biographical sketch} (3041), followed by
\textit{Funeral information} (2831) and \textit{Family} (2195). The least
assigned label is \textit{Tribute}, with 11 sentences, followed by
\textit{Gratitude} with 144 sentences.

Sentences of class Biographical Sketch and Characteristics are more frequent in
obituaries from the US than from Canada and UK. On the other side, Family is a
more dominant class ins UK than in the other sources.

Surprisingly, the class \textit{Funeral information} is also not
equally distributed across locations, which is dominated by the UK.

Finally, Canada has a substantially higher section of sentences
labeled with \textit{Other}. A manual inspection of the annotation
showed that this is mostly because it seems to be more common than in
other locations to mention that the person will be remembered.

\section{Methods}\label{sec:methods}
\begin{table*}
  \centering
  \begin{tabular}{lcccccccccccc}
    \toprule
    &  \multicolumn{3}{c}{CNN}& \multicolumn{3}{c}{BiLSTM (BOW)} & \multicolumn{3}{c}{BiLSTM (W2V)} & \multicolumn{3}{c}{BiLSTM-CRF}\\
      \cmidrule(lr){2-4}\cmidrule(lr){5-7}\cmidrule(lr){8-10}\cmidrule(lr){11-13}
    Class &P&R&F1&P&R&F1&P&R&F1&P&R&F1\\
   \cmidrule(r){1-1}\cmidrule(lr){2-4}\cmidrule(lr){5-7}\cmidrule(lr){8-10}\cmidrule(lr){11-13}
    Personal Information & .88 & .85 & .83 &
    .89 & .85 & .87 &
    .90 & .82 & .86 &
    .96 & .78 & .86 \\
    Biographical Sketch & .82 & .80 & .81 &
    .80 & .89 & .84 &
    .78 & .64 & .70 &
    .76 & .56 & .64\\
    Family & .92 & .89 & .90 &
    .86 & .89 & .87 &
    .64 & .90 & .74 &
    .85 & .85 & .85 \\
    Characteristics & .65 & .75 & .70 &
    .53 & .73 & .61 &
    .60 & .51 & .56 &
    .38 & .75 & .50\\
    Tribute & .00 & .00& .00& .00& .00& .00& .00& .00& .00& .00 & .00 & .00 \\
    Gratitude & .75 & .47 & .57 &
    .60 & .20 & .30 &
    .76 & .18 & .29 &
    .94 & .37 & .53 \\
    Funeral Information & .84 & .97 & .90 &
    .86 & .98 & .92 &
    .74 & .99 & .85 &
    .77 & .99 & .87\\
    Other & .59 & .46 & .52 &
    .52 & .11 & .18 &
    .67 & .01 & .03 &
    .53 & .13 & .21 \\
    \cmidrule(r){1-1}\cmidrule(lr){2-4}\cmidrule(lr){5-7}\cmidrule(lr){8-10}\cmidrule(r){11-13}
    Macro  & {.68} & {.64} & {.65}& {.63} & {.58} & {.58} & {.63} & {.50} & {.50} & {.64} & {.55} & {.50}\\
    Micro  & & & {.81} & & & {.80} & & & {.72} & & & {.73}\\
    \bottomrule
  \end{tabular}
  \caption{Comparison of the models using Precision, Recall, and F1-score (macro and micro)}
  \label{tab:results}
\end{table*}

To answer the question whether or not we can recognize the structure in
obituaries we formulate the task as sentence classification, where each
sentence will be assigned to one of the eight classes we defined previously. We
evaluate four different models.

\paragraph{CNN} Convolutional Neural Networks (CNN)
\cite{collobert2008,kim2014} have been succesfully applied to practical NLP
problems in the recent years. We use the sequential model in
Keras\footnote{\url{https://keras.io/}} where each sentence is represented as a
sequence of one-hot embeddings of its words.
We use three consecutive pairs of convolutional layers with 128 output
channels, the ReLu activation function and max pooling followed by the output
layer with softmax as activation function and with cross entropy as loss. This
model does not have access to information of neighboring sentences.

\paragraph{BiLSTM}
The BiLSTM models are structurally different from the CNN. The CNN predicts on
the sentence-level without having access to neighboring information. For the
BiLSTM models we opt for a token-based IOB scheme in which we map the
dominantly predicted class inside of one sentence to the whole sentence. Our
\textbf{BiLSTM (BOW)} model \cite{lstm,Schuster1997} uses 100 memory units, a
softmax activation function and categorical cross entropy as the loss function.
The \textbf{BiLSTM (W2V)} model uses pre-trained word embeddings (Word2Vec on
Google News) \cite{Mikolov13} instead of the bag of words. The
\textbf{BiLSTM-CRF} is an extension of the \textbf{BiLSTM (W2V)} which uses a
conditional random field layer for the output.

\section{Experimental Setup}\label{sec:exp}

\begin{figure}
  \centering
  \includegraphics[width=0.99\columnwidth]{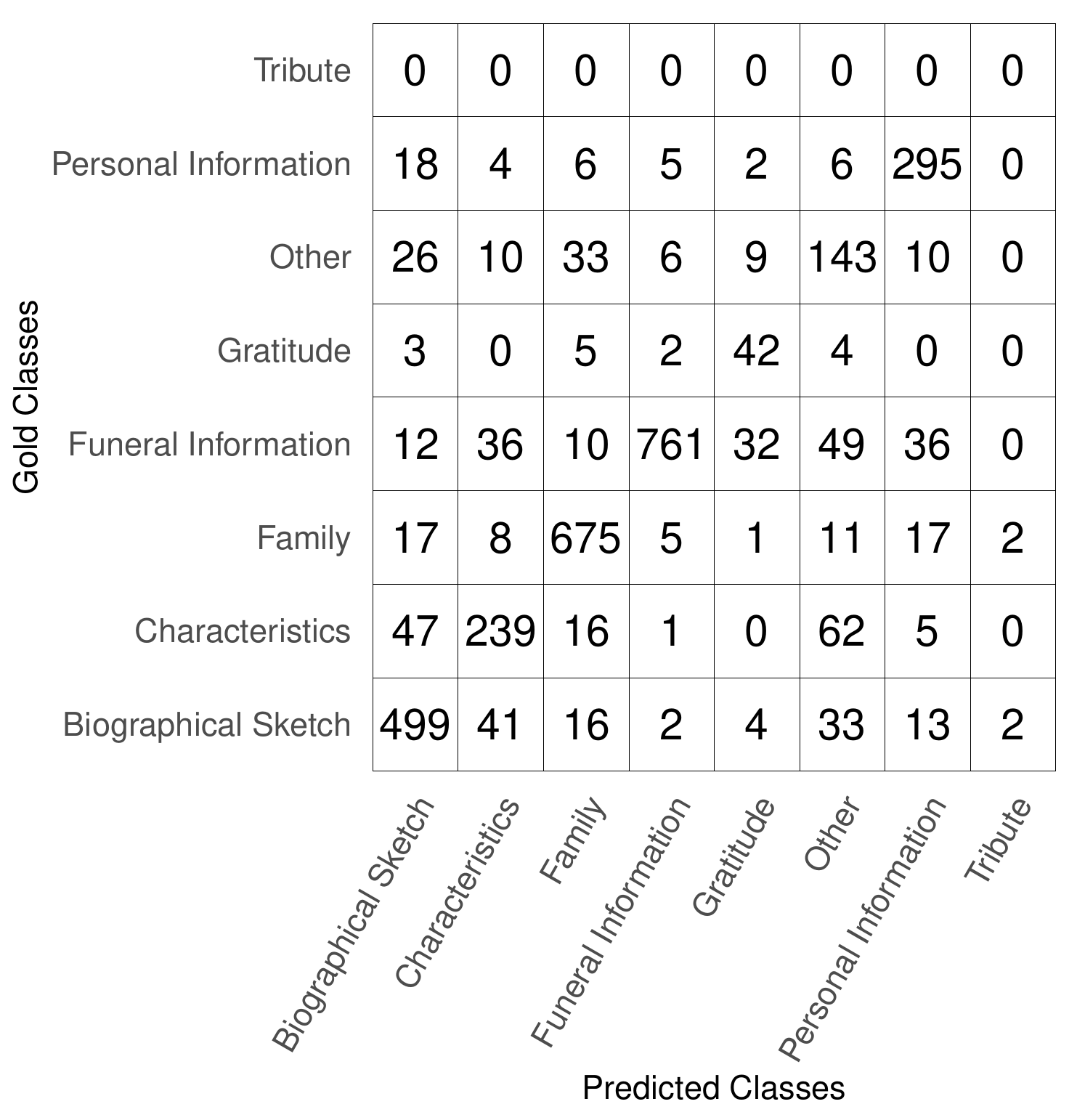}
  \caption{Confusion matrix of the CNN model.}
  \label{Table_7}
\end{figure}

\begin{table*}
 \centering
 \footnotesize
  \begin{tabularx}{\linewidth}{p{0.2cm}p{0.6cm}p{0.6cm}p{12cm}p{1.6cm}}
    \toprule
    \# & Gold & Pred.  & Sentence & Error Type \\
    \cmidrule(rl){1-1}\cmidrule(rl){2-2}\cmidrule(rl){3-3}\cmidrule(rl){4-4}\cmidrule(l){5-5}
    1 & F & BS &	John passed away in 2001 prior to Mary's retirement. & Ambiguity \\
    2 & F & PI & Passed away in Vancouver, British Columbia on November 10, 2018. & Annotation\\
    \cmidrule(rl){1-1}\cmidrule(rl){2-2}\cmidrule(rl){3-3}\cmidrule(l){3-3}\cmidrule(l){4-4}\cmidrule(l){5-5}
    3 & C &	O &	We will all miss his wit and his incredible sense of humour. & Other \\
    4 & C & BS &	Together they shared golf and travel as well as the Legion. & Ambiguity \\
     \cmidrule(rl){1-1}\cmidrule(rl){2-2}\cmidrule(rl){3-3}\cmidrule(l){3-3}\cmidrule(l){4-4}\cmidrule(l){5-5}
 5 & FI & O	& May you find comfort in the arms of angels.  & Ambiguity \\
   6 & FI &	G &	The family would like to thank Dr. J. Doe and  the amazing staff at the Cross Cancer Institute for their compassion and excellent care of our Opa and our family.  & Annotation \\
      \cmidrule(rl){1-1}\cmidrule(rl){2-2}\cmidrule(rl){3-3}\cmidrule(l){3-3}\cmidrule(l){4-4}\cmidrule(l){5-5}
 7 & BS &	C &	Christian radio broadcasts led him to his Savior Jesus Christ. & Ambiguity \\
 8 & BS &	F & Mary was born on September 5, 1908 in Edam, Saskatchewan and was one of eight siblings.& Ambiguity \\
 9 & BS & C &	John was an accomplished musician. & Annotation \\
 10 & BS & C &	Sadly Mary had a stroke and John became his primary care giver where she did an exceptional job. & Other \\
    \cmidrule(rl){1-1}\cmidrule(rl){2-2}\cmidrule(rl){3-3}\cmidrule(l){3-3}\cmidrule(l){4-4}\cmidrule(l){5-5}
 11 & O &	BS &	John was successful in business and in life having many friends in both. & Other \\
 12 & O & C &	To say he was well-liked would be an understatement, he was well-loved. & Annotation \\
13 & O	& F & This treatment resulted in the survival of countless otherwise terminally ill children & Ambiguity \\
14 & O & FI &	You are truly special & Ambiguity \\
15 & O & C &	It was a joy and privilege to share our lives with you. & Other \\
    \cmidrule(rl){1-1}\cmidrule(rl){2-2}\cmidrule(rl){3-3}\cmidrule(l){3-3}\cmidrule(l){4-4}\cmidrule(l){5-5}
16 & PI	& F & FOREVER IN OUR HEARTS. Lovingly remembered by Mary (Bob), Anne (Colleen), Laura (Alice, Oscar, Jen) & Annotation \\
17 & PI &	BS & 	In spite of having cancer, she was brave and determined to enjoy life to the fullest throughout this past year. & Other \\
     \cmidrule(rl){1-1}\cmidrule(rl){2-2}\cmidrule(rl){3-3}\cmidrule(l){3-3}\cmidrule(l){4-4}\cmidrule(l){5-5}
18 & T &	C &	His work was a credit to the Ukrainian community, elevating the efforts of its arts sector beyond its own expectations.  & Other \\
19 & T & 	F  &	John's awards, professional designations and charitable associations are too great to list but to us he was our much loved husband, father, brother, grandfather and we will miss him. & Ambiguity \\
     \cmidrule(rl){1-1}\cmidrule(rl){2-2}\cmidrule(rl){3-3}\cmidrule(l){3-3}\cmidrule(l){4-4}\cmidrule(l){5-5}
20 & G	& FI &	Mary's family would like to thank the care team at Extendicare Holyrood for their kind care and attention in making Mary comfortable, happy, and safe. & Other \\
21 & G	& F &	Thanks are also extended to the many friends and family who have been there, your love and support is immeasurable and thank you will never be enough. & Other \\
     \bottomrule
  \end{tabularx}
  \caption{Example of errors done by the CNN model. PI: Personal information, BS: Biographical Sketch, FA: Family, C:
    Characteristics, T: Tribute, G: Gratitude, FI: Funeral
    Information, O: Other}  \label{obituary_errors}
\end{table*}

We split our 1008 obituaries into training set (70\,\%) and test set (30\,\%).
From the training set, 10\,\% are used for validation. The batch size is set to
8 and the optimizer to \textit{\hbox{rmsprop}} for all experiments. We do not perform
hyperparameter tuning.

\subsection{Results}
\label{sec:results}

The CNN model has the highest macro average \F score with a value of 0.65. This
results from the high values for the classes \textit{Family} and
\textit{Funeral information}. The \F score for the class \textit{Other} is 0.52
in contrast with the \F of the other three models, which is lower than 0.22.
The macro average \F for the BiLSTM (BOW) model is 0.58. It also has highest
F1-scores for the classes \textit{Personal Information} and
\textit{Biographical Sketch} among all models. For the classes \textit{Family},
and \textit{Funeral information} has comparable scores to the CNN model.
Interestingly this model performs the best among the BiLSTM variants.  The
BiLSTM (W2V) model performs overall worse than the one which makes use only of
a BOW. It also has the worst macro average \F together with the BiLSTM-CRF with a value of 0.50. The BiLSTM-CRF performs better than the other
BiLSTM variants on the rare classes \textit{Gratitude} and \textit{Other}.

Since we have few samples labelled as \textit{Tribute} none of our models
predict a sentence as such, resulting in precision, recall, and \F value of 0
for each model.

From the results we conclude that the CNN model works best. Apart from the high
\F it is also the only model that predicts the class \textit{Gratitude} as well
as the class \textit{Other} better than the other models.

\subsection{Error Analysis}
We investigate the best performing model by making use of the confusion matrix
(see Figure~\ref{Table_7}) and by inspecting all errors made by the model on
the test set (see Table~\ref{obituary_errors}).

In Figure~\ref{Table_7}, we observe that the diagonal has relatively high numbers with
more correctly labeled instances than confused ones for all classes, with the
exception of class \textit{Tribute} (the rarest class). Secondly, the
confusions are not globally symmetric. However, we observe that the lower left
corner formed by the classes \textit{Family}, \textit{Characteristics} and
\textit{Biographical Sketch} is almost symmetric in its confusions, which led
us to inspect and classify the types of errors.

Therefore, we investigated all errors manually and classified them in three
main types of errors: errors due to \textit{Ambiguity} (39\%), errors due to
wrong \textit{Annotation} (18\%) and errors tagged as \textit{Other} (42\%)
where the errors are more difficult to explain (see last column in
Table~\ref{obituary_errors}).

The errors due to \textit{Ambiguity} are those where a test sentence could be
reasonably assigned multiple different zones, and both the annotated class and
the predicted class would be valid zones of the sentence. Such cases are most
common between the zones \textit{Biographical Sketch}, \textit{Personal
Information}, \textit{Characteristics}, \textit{Other}, and \textit{Family} and
occur even for the rare zones \textit{Tribute} and \textit{Gratitude}. An
example of this error type is sentence 7 in Table~\ref{obituary_errors}, which
shows that there is a significant event that happened in the life of the
deceased that changed their characteristics.

Another pattern we observe emerging within the \textit{Ambiguity} class of
errors is that borders between the classes confused are not as rigid, and
sometimes parts of one class could be entailed in another. An example of this
is when the class \textit{Other} being entailed in \textit{Funeral Information}
or \textit{Characteristics} as a quote,  as a wish in sentence 5 (\eg, ``may
you find comfort...'') or as a last message from the family to the deceased
(\eg ``You are truly special.'') in sentence 14.

The errors we mark as being errors of \textit{Annotation} are those where the
model is actually right in its prediction. Such cases are spread among all
classes. The class that is the most affected by these errors is the
class \textit{Characteristics}, for which there are $23$ cases of sentences
wrongly annotated as being in the class \textit{Other} or \textit{Biographical
Sketch} (\eg sentences 9, 12). The second most affected class by this type of
error is \textit{Biographical Sketch} where the sentences are also wrongly
annotated as \textit{Other}. The rare class \textit{Gratitude} is also $13$
time wrongly annotated as \textit{Other}, \textit{Personal Information} or
\textit{Biographical Sketch}. This might explain why the model confuses these
classes as well (Figure~\ref{Table_7}) \ Other examples for this type of error
we can see for sentence 2, 6 and  16.

The rest of the errors, labeled here as \textit{Other}, are diverse and more
difficult to categorize. However, we see a pattern within this group of errors
as well, such as when the model appears to be mislead by the presence of words
that are strong predictive features of other classes. This could be seen for
instance in sentence 19 where \textit{Gratitude} in confused with
\textit{Family} due to the presence of words like ``family'', ``love'',
``support''. This type of error can be also seen in sentence 11, 19. Another
pattern that shows for errors of the type \textit{Other} is when the model
fails to predict the correct class because is not able to do coreference
resolution as in sentences 10 and 15.

Regarding \textit{Gratitude}, the confusion matrix shows that it is confounded with
\textit{Family}, \textit{Other}, and \textit{Funeral Information}. Inspecting
these cases shows that the wrongly classified cases are due to the presence of
strong predictive features of other classes, like family mentions or locations
which are more prevalent in other classes as in the sentences 18 and 19.

Further, the class \textit{Funeral Information} is confounded the most with
\textit{Other}, followed by \textit{Personal Information} and
\textit{Characteristics}. We see a high number of confusions between
\textit{Funeral Information} and \textit{Gratitude} as well, and since
\textit{Gratitude} is one of the rare classes we decide to have a closer look
at these cases. We find that most of the misclassified sentences include
expressions of gratitude and are therefore wrongly annotated, which shows that
the model correctly learned that expressions like ``would like to thank'',
``thanks'', ``thank you'' include predictive features for the class
\textit{Gratitude} (see sentence 6).

When the class \textit{Characteristics} is confounded with \textit{Other}, this
happens mostly due to presence of words related to memory ``we will miss'',
``we will always remember'', ``our memories'', ``will be deeply missed'' which
are most occurring within the class \textit{Other}. This hints to a potential
improvement in the Annotation Scheme, where one could add the class
\textit{Societal Memory} where all the sentences that mention what the
community will miss due to the loss would belong to. We think that another
improvement would be if the class \textit{Other} would be further divided into
\textit{Wish} and \textit{Quote} as well, this would eliminate the issue of
entailed sentences of \textit{Other} in other classes.

\section{Conclusion and Future Work}

This work addresses the question of how to automatically structure obituaries.
Therefore, we acquire a new corpus consisting of 20058 obituaries of which 1008
are annotated. To tackle the task of assigning zones to sentences and uncover
the structure of obituaries, four segmentation models are implemented and
tested: a CNN, a BiLSTM network using a BOW model and one using word
embeddings, and a BiLSTM-CRF. The models are then compared based on precision,
recall, and F1-score. From our results, we conclude that the CNN text
classifier produced the best results with a macro F1-score of 0.81, considering
the experimental settings, and the highest macro average F1-score of 0.65. The
BiLSTM (BOW) model produced comparable results and even better regarding the
classes \textit{Personal information} and \textit{Biographical sketch}, which
makes it also a valid baseline for the task.

Our work enables future research, showing that automatic recognition of
structures in obituaries is a viable task. Through performing zoning on the raw
obituaries, it is becoming possible to address other research questions:
whether there is a correlation between the occupation of the deceased and the
cause of death, what are the cultural and structural differences between
obituaries from different countries.

Another open question is if the annotation scheme is the best. Given the errors
we found, we argue that the annotation scheme could be refined and that the
class \textit{Other} could be split into at least two different new classes. We
leave to future work developing a new annotation scheme. Further, one could
annotate obituaries across cultures, optimize the parameters of our models for
the structuring task or improve over the existing models. It might be an
interesting direction to compare our defined structure with one of a topic
modeler. Also possible is to postannotate the dataset with emotion classes and
investigate the emotional connotation of different zones.

\section{Bibliographical References}
\label{main:ref}

\end{document}